\documentclass[conference]{IEEEtran}
\IEEEoverridecommandlockouts
\usepackage{cite}
\usepackage{amsmath,amssymb,amsfonts}
\usepackage{algorithmic}
\usepackage{algorithm}
\usepackage{graphicx}
\usepackage{textcomp}
\usepackage{xcolor}
\usepackage{array}
\usepackage{verbatim}
\usepackage{multirow}
\usepackage{bm}
\usepackage{hyperref}
\usepackage{url}
\usepackage[caption=false,font=footnotesize,labelfont=rm,textfont=rm]{subfig}
\usepackage[framemethod=tikz]{mdframed}
\usepackage{color,soul}
\usepackage{balance}

\def\BibTeX{{\rm B\kern-.05em{\sc i\kern-.025em b}\kern-.08em
    T\kern-.1667em\lower.7ex\hbox{E}\kern-.125emX}}
\begin{document}

\title{Towards Feature Distribution Alignment and Diversity Enhancement for Data-Free Quantization}

\author{
\IEEEauthorblockA{Yangcheng~Gao$^{1,2}$, Zhao~Zhang$^{1,2}$, Richang~Hong$^{1,2}$, Haijun~Zhang$^{3}$, Jicong~Fan$^{4,5}$, Shuicheng~Yan$^{6}$}
$^{1}$ School of Computer Science and Information Engineering, Hefei University of Technology, Hefei 230009, China\\
$^{2}$ Key Laboratory of Knowledge Engineering with Big Data (Ministry of Education) \& Intelligent Interconnected Systems\\
Laboratory of Anhui Province, Hefei University of Technology, Hefei 230009, China\\
$^{3}$ Department of Computer Science, Harbin Institute of Technology (Shenzhen), Xili University Town, Shenzhen, China\\
$^{4}$ School of Data Science, The Chinese University of Hong Kong, Shenzhen, China \\
$^{5}$ Shenzhen Research Institute of Big Data, Shenzhen, China \\
$^{6}$ National University of Singapore, Singapore 117583\\
E-mails: gaoyangcheng576@gmail.com, cszzhang@gmail.com
}

\maketitle

\begin{abstract}

To obtain lower inference latency and less memory footprint of deep neural networks, model quantization has been widely employed in deep model deployment, by converting the floating points to low-precision integers. However, previous methods (such as quantization aware training and post training quantization) require original data for the fine-tuning or calibration of quantized model, which makes them inapplicable to the cases that original data are not accessed due to privacy or security. This gives birth to the data-free quantization method with synthetic data generation. While current data-free quantization methods still suffer from severe performance degradation when quantizing a model into lower bit, caused by the low inter-class separability of semantic features. To this end, we propose a new and effective data-free quantization method termed ClusterQ, which utilizes the feature distribution alignment for synthetic data generation. To obtain high inter-class separability of semantic features, we cluster and align the feature distribution statistics to imitate the distribution of real data, so that the performance degradation is alleviated. Moreover, we incorporate the diversity enhancement to solve class-wise mode collapse. We also employ the exponential moving average to update the centroid of each cluster for further feature distribution improvement. Extensive experiments based on different deep models (e.g., ResNet-18 and MobileNet-V2) over the ImageNet dataset demonstrate that our proposed ClusterQ model obtains state-of-the-art performance.

\end{abstract}


\section{Introduction}
\label{sec:intro}

Deep neural network (DNN)-based models have obtained remarkable progress on computer vision tasks due to its strong representation ability 
\cite{krizhevsky2012imagenet,he2016deep,szegedy2015going,wei2021deraincyclegan,ren2021robust,zhang2021discriminative,zhang2021dual,ji2021cnn}. However, DNN models usually suffer from high computational complexity and massive parameters, and large DNN models require frequent memory access, which will lead to much more energy consumption and inference latency \cite{han2016eie}. Moreover, it is still challenging to deploy them on the edge devices due to the limited memory bandwidth, inference ability and energy consumption.

To address aforementioned issues, massive model compression methods have emerged to improve the efficiency of DNN models, such as pruning \cite{liu2018rethinking,ruan2020edp,chen2020dynamical}, quantization \cite{jacob2018quantization,banner2019post, choi2018pact, courbariaux2015binaryconnect,Esser2020LEARNED,kim2020exploiting,cai2020zeroq,choi2021qimera,xu2020generative,zhong2021fine,nahshan2021loss,zhu2021autorecon,zhang2021diversifying}, light-weight architecture design \cite{sandler2018mobilenetv2,howard2019searching,ma2018shufflenet}, low-rank factorization \cite{denton2014exploiting,jaderberg2014speeding,9679094} and knowledge distillation \cite{cheng2020explaining,wang2020real}. Different from other model compression methods, model quantization can be implemented in real-scenario model deployment, with the low-precision computation supported on general hardware. Briefly, model quantization paradigm converts the floating-point values into low-bit integers for model compression \cite{jacob2018quantization}. As such, less memory access will be needed and the computational latency will be reduced in model inference, which will make it possible to deploy large DNN models on the edge devices for those real-time applications.

Due to the limited representation ability over low-bit values, model quantization usually involves noise, which potentially results in the performance degradation in reality. To recover the quantized model performance, Quantization Aware Training (QAT) performs backward propagation to retrain the quantized model \cite{courbariaux2015binaryconnect,choi2018pact,Esser2020LEARNED,kim2020exploiting}. However, QAT is usually time-consuming and hard to implement, so Post Training Quantization (PTQ), as an alternative method, aims at adjusting the weights of quantized model without training \cite{banner2019post,nahshan2021loss,zhong2021fine}. Note that QAT and PTQ need the original training data for quantization, whereas training data may be prohibited severely from access due to privacy or proprietary rules in real scenario, e.g., user data, military information, or medical images. As a result, real-world applications of QAT and PTQ may be restricted.

\begin{figure}[!t]
	\centering
	\subfloat[]{\includegraphics[width=0.75\linewidth]{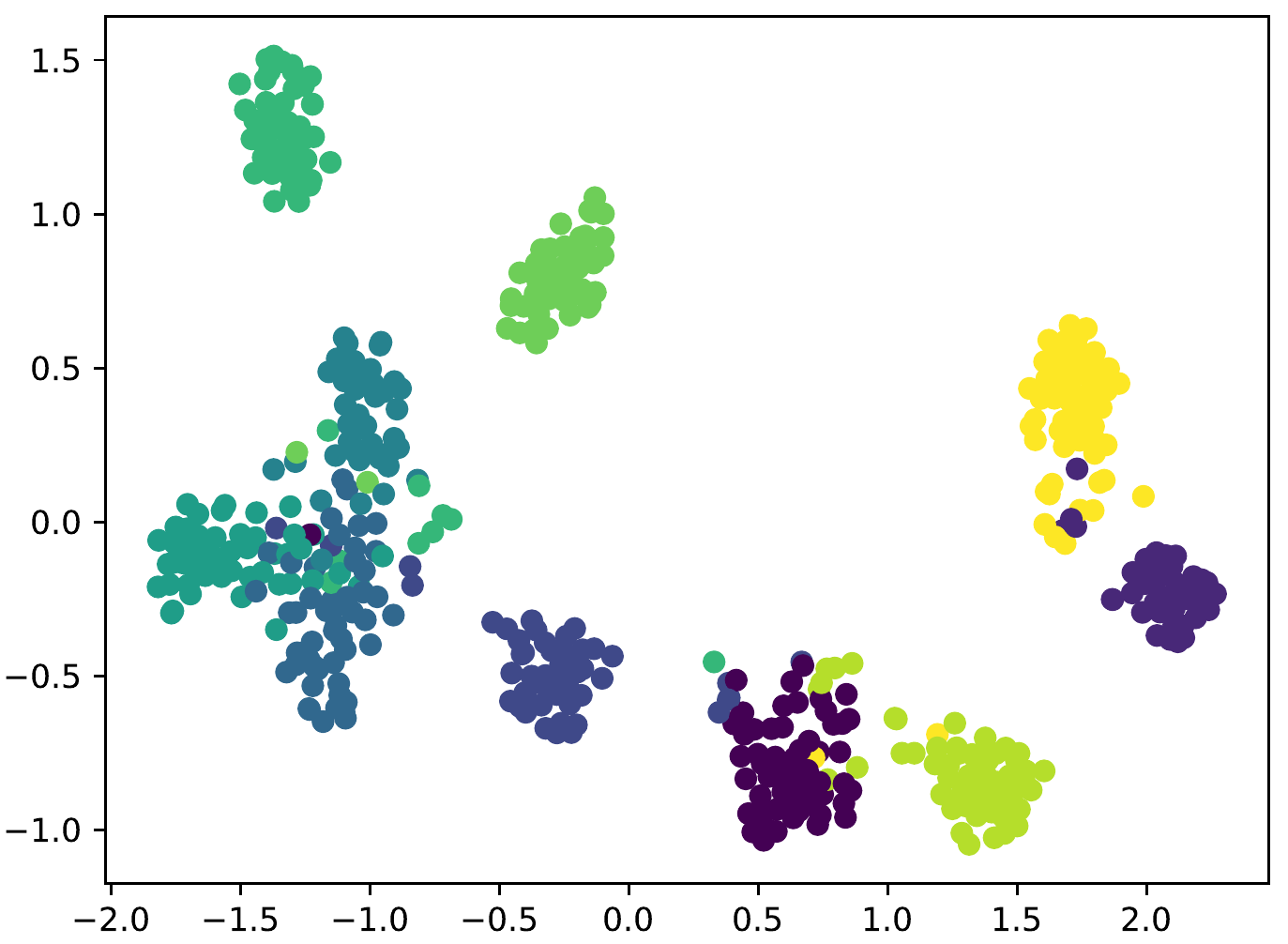}}
	\quad
	\subfloat[]{\includegraphics[width=0.75\linewidth]{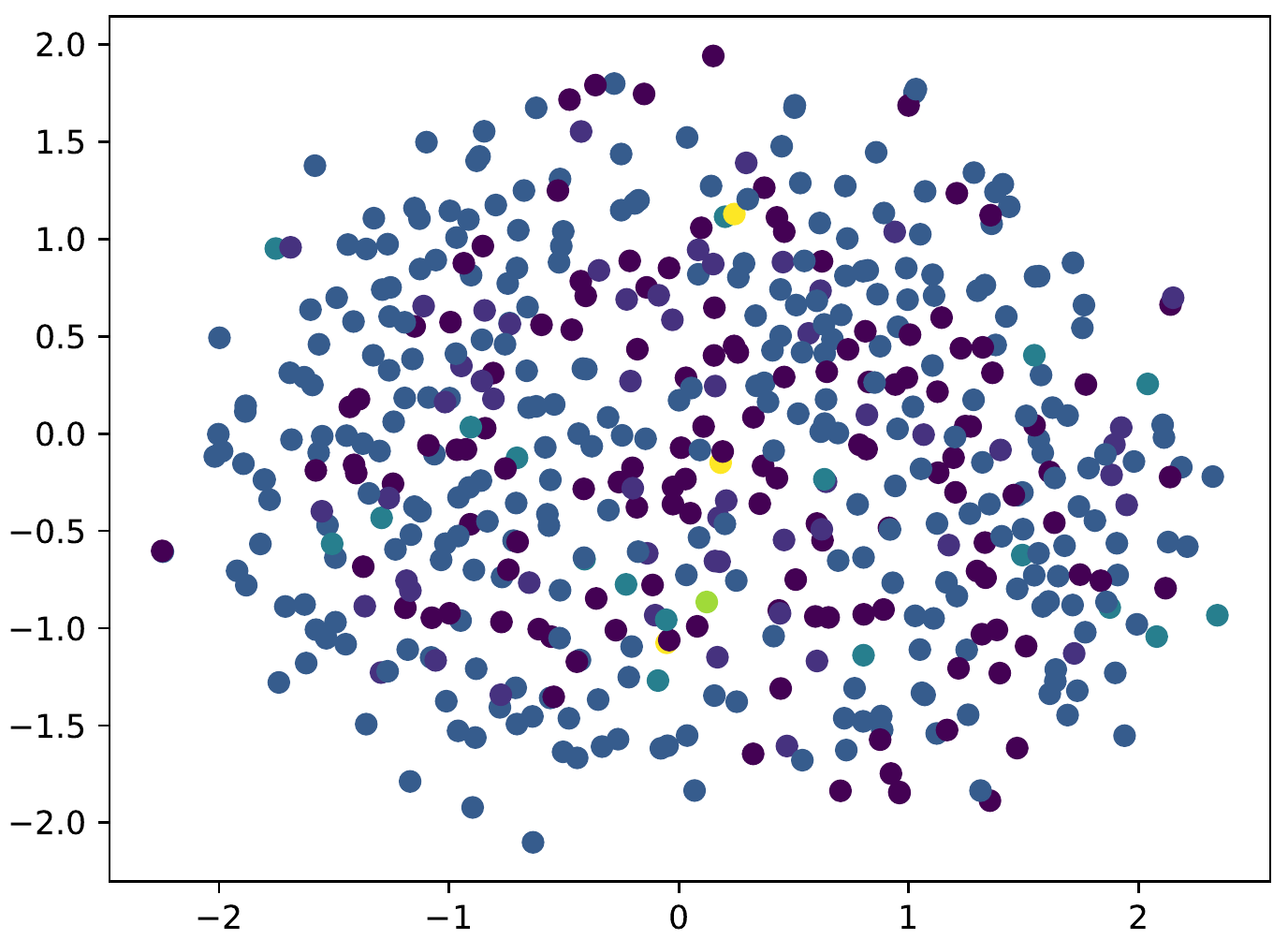}}
	\caption{Visualization of the feature distribution of ResNet-20 \cite{he2016deep} running on the CIFAR10 \cite{krizhevsky2009learning} (a), and the synthetic data generated by ZeroQ \cite{cai2020zeroq} (b). Clearly, ZeroQ cannot produce features with high inter-class separability.}
	\label{fig:compare}
\end{figure}

Recently, Data-Free Quantization (DFQ) have came into being as a more promising method for practical applications without access to any training data, which aims at restoring the performance of quantized model by generating synthesis data, similar to the data-free knowledge distillation \cite{lopes2017data}. Current DFQ methods can be roughly divided into two types, i.e., without fine-tuning and with fine-tuning. Pioneer work on DFQ without fine-tuning, like ZeroQ \cite{cai2020zeroq}, generate the calibration data that matches the batch normalization (BN) statistics of model to clip the range of activation values. However, compressed models by this way often have significant reduction in accuracy when quantizing to lower precision. In contrast, DFQ with fine-tuning applies generator to produce synthetic data and adjusts the parameters of quantized model to retain higher performance. For example, Generative Low-bitwidth Data Free Quantization (GDFQ) \cite{xu2020generative} learns a classification boundary and generates data with a Conditional Generative Adversarial Network (CGAN)  \cite{mirza2014conditional}. 

Although recent studies have witnessed lots of efforts on the topic of DFQ, the obtained improvements are still limited compared with PTQ, due to the gap between the synthetic data and real-world data. As such, how to make the generated synthetic data closer to real-world data for fine-tuning will be a crucial issue to be solved. To close the gap, we explore the pre-trained model information at a fine-grained level. According to \cite{li2016revisiting,wan2018rethinking}, during the DNN model inferring on real data, the distributions of semantic features can be clustered for classification, i.e., inter-class separability property of semantic features. This property has also widely used in domain adaption to align the distributions of different domains. However, the synthetic data generated by current DFQ methods (such as ZeroQ \cite{cai2020zeroq}) still cannot produce features with high inter-class separability in the quantized model, as shown in Figure \ref{fig:compare}. Based on this phenomenon, we can hypothesize that high inter-class separability will reduce the gap between synthetic data and real-world data. Note that this property has also been explored by FDDA \cite{zhong2021fine}, which augments the calibration dataset of real data for PTQ. However, there still does not exist data-free quantization method that imitates the real data distribution with inter-class separability.

\begin{figure*}[t]
	\makeatletter
	\makeatother
	\centering
	\includegraphics[width=0.88\linewidth]{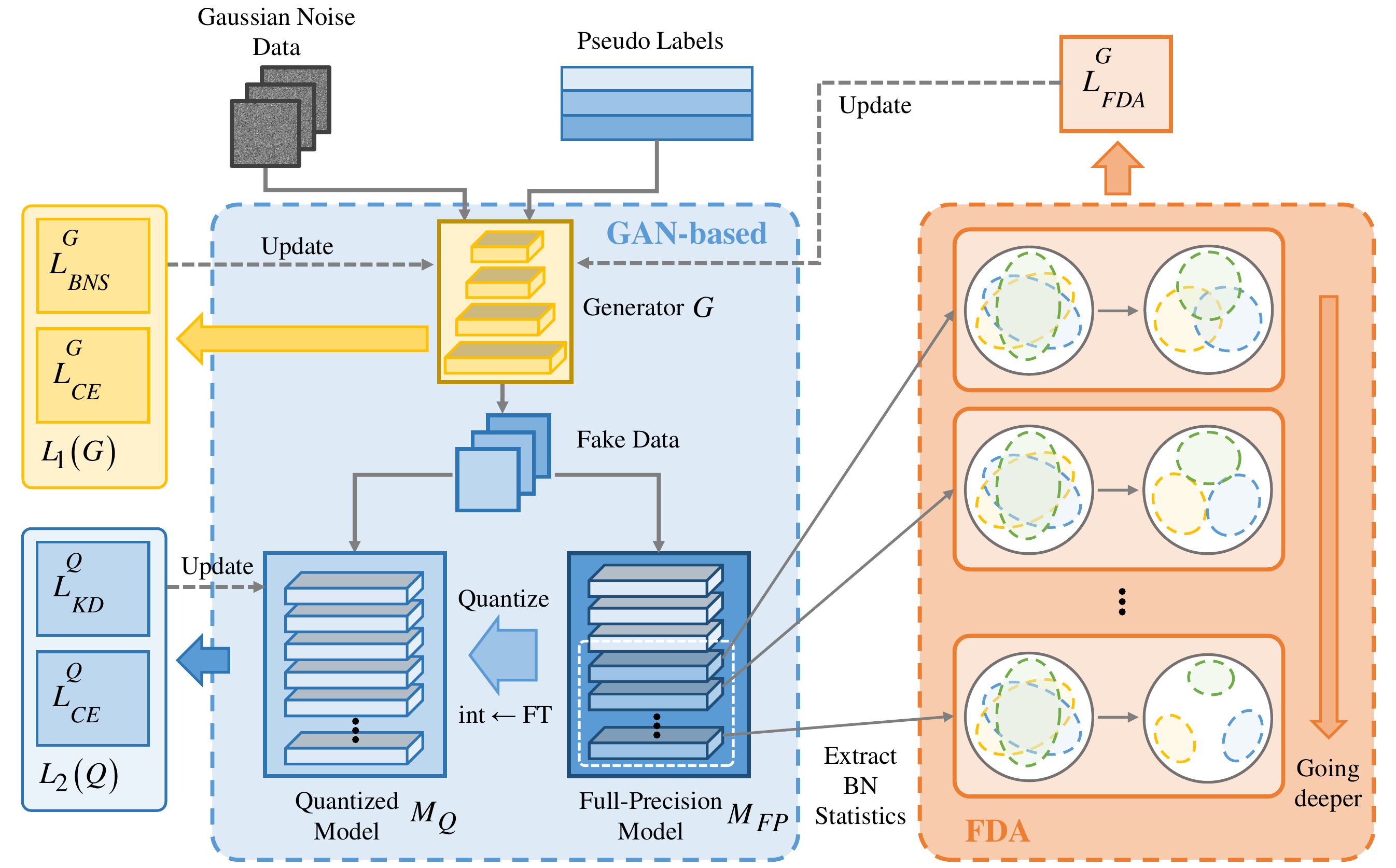}
	\caption{Overview of the proposed ClusterQ scheme. Based on the Conditional Generative Adversarial Network (CGAN) \protect\cite{mirza2014conditional} mechanism, we perform clustering and alignment on the batch normalization statistics of semantic features to obtain high inter-class separability.}
	\label{fig:overview}
	\vspace{-2mm}
\end{figure*}

From this perspective, we will propose effective strategies to generate synthetic data to obtain features with high inter-class separability and maintain the generalization performance of the quantized model for data-free case. In summary, the major contributions of this paper are described as follows: 

\begin{enumerate}
    \item Technically, we propose a new and effective data-free quantization scheme, termed ClusterQ, via feature distribution clustering and alignment, as shown in Figure \ref{fig:overview}. As can be seen, ClusterQ formulates the DFQ problem as a data-free domain adaption task to imitate the distribution of original data. To the best of our knowledge, ClusterQ is the first DFQ scheme to utilize feature distribution alignment with clusters. 
    \item This study also reveals that high inter-class separability of the semantic features is critical for synthetic data generation, which impacts the quantized model performance directly. We quantize and fine-tune the DNN model with a novel synthetic data generation approach without any access to the original data. To achieve high inter-class separability, we propose a Feature Distribution Alignment (FDA) method, which can cluster and align the semantic feature distribution to the centroids for close-to-reality data generation. For further performance improvement, we introduce the diversity enhancement process to enhance the data diversity and exponential moving average (EMA) to update the cluster centroids. 
    \item Based on the clustered and aligned semantic feature distributions, our ClusterQ can effectively alleviate the performance degradation, and obtain state-of-the-art results on a variety of popular deep models. 
\end{enumerate}

The rest of this paper is organized as follows. In Section \ref{sec:relate}, we review the related work. The details of our method are elaborated in Section \ref{sec:method}. In Section \ref{sec:exp}, we present experiment results and analysis. The conclusion and perspective on future work are finally discussed in Section \ref{sec:conc}.

\section{Related Work}
\label{sec:relate}

We briefly review the low-bit quantization methods that are close to our study. More details can be referred to \cite{gholamisurvey} that provides a comprehensive overview for model quantization.

\subsection{Quantization Aware Training (QAT)}

To avoid performance degradation of the quantized model, QAT is firstly proposed to retrain the quantized model \cite{courbariaux2015binaryconnect,choi2018pact,Esser2020LEARNED,kim2020exploiting}. With full training dataset, QAT performs floating-point forward and backward propagations on DNN models and quantizes them into low-bit after each training epoch. Thus, QAT can quantize model into extremely low precision while retaining the performance. In particular, PACT \cite{choi2018pact} optimizes the clipping ranges of activations during model retraining. LSQ \cite{Esser2020LEARNED} learns step size as a model parameter and MPQ \cite{kim2020exploiting} exploits retraining-based mix-precision quantization. However, high computational complexity of QAT will lead to restrictions on the implementation in reality. 

\subsection{Post Training Quantization (PTQ)}

PTQ is proposed for efficient quantization \cite{banner2019post,nahshan2021loss,zhong2021fine}. Requiring for a small amount of training data and less computation, PTQ methods have ability to quantize models into low-bit precision with little performance degradation. In particular, \cite{banner2019post} propose a clipping range optimization method with bias-correction and channel-wise bit-allocation for 4-bit quantization. \cite{nahshan2021loss} explore the interactions between layers and propose layer-wise 4-bit quantization. \cite{zhong2021fine} explore calibration dataset with synthetic data for PTQ.  However, above methods require more or less original training data, and they are inapplicable for the cases without access to original data. 

\subsection{Data-Free Quantization (DFQ)}

For the case without original data, recent studies made great efforts on DFQ to generate the close-to-reality data for model fine-tuning or calibration
\cite{cai2020zeroq,xu2020generative,choi2021qimera,zhang2021diversifying,zhu2021autorecon}. Current DFQ methods can be roughly divided into two categories, i.e., without fine-tuning and with fine-tuning. Pioneer work on DFQ without fine-tuning, like ZeroQ \cite{cai2020zeroq}, generate the calibration data that matches the batch normalization (BN) statistics. DSG \cite{zhang2021diversifying} discovers homogenization of synthetic data and enhances the diversity of generated data. However, these methods lead to significant reduction in accuracy when quantizing to lower precision. In contrast, DFQ with fine-tuning applies generator to produce synthetic data and adjusts the parameters of quantized model to retain higher performance. For example, GDFQ \cite{xu2020generative} employs a Conditional Generative Adversarial Network (CGAN) \protect\cite{mirza2014conditional} mechanism and generates dataset for fine-tuning. AutoReCon \cite{zhu2021autorecon} enhances the generator by neural architecture search. Qimera \cite{choi2021qimera} exploits boundary supporting samples to enhance the classification boundary. In addition, DFQ method without data generation has also emerged \cite{guo2022squant} which will lead to higher inference latency than generative methods due to the dynamic range clipping.

\section{Methodology}
\label{sec:method}

We will firstly review some preliminaries. Then, the design of feature distribution alignment will be presented in detail. We will also introduce the center updating and diversity enhancement. Finally, we show a comprehensive description for the proposed framework at the end of this section. 

\subsection{Preliminary}

\subsubsection{Batch Normalization}
Since we exploit the Batch Normalization (BN) statistics to imitate the original distribution, we first review the BN layer briefly \cite{ioffe2015batch}, which is designed to reduce the internal covariate shifting. Formally, with a mini-batch input $\bm{X}_B= \left\{ \bm{x}_1, \bm{x}_2, ... , \bm{x}_m \right\}$ of batch size $m$, the BN layer will transfer the input $\bm{X}_B$ into the following expression:

\begin{equation}
\begin{aligned}
    \bm{\hat x}_i &\xleftarrow{} \frac{\bm{x}_{i} - E[\bm{X}_B]}{\sqrt{Var[\bm{X}_B] + \epsilon}},\\
    \bm{y}_i &\xleftarrow{} \gamma_i \bm{\hat x}_i + \beta_i,
\end{aligned}
\end{equation}

where $x_i$ and $y_i$ denote the input and output of BN layer respectively, $\gamma_i$ and $\beta_i$ denote the parameters learned during training. In addition, mean and standard deviation parameters, i.e. $\mu$ and $\sigma$, are stored in each BN layer and are used to describe the feature distribution. 

\subsubsection{Model Quantization}
For easy implementation on hardware, our ClusterQ employs a symmetric uniform quantization, which maps and rounds the floating points of full-precision model to low-bit integers. Given a floating-point value $x$ in a tensor $\bm{x}$ to be quantized, it can be defined as follows: 
\begin{equation}
    \hat x =round(\frac{x}{S}),\  S=\frac{2\alpha}{2^N-1},
\label{eq:quant}
\end{equation}
where $N$ denotes the bit width for quantizing, $\alpha$ denotes the clipping range for floating points, $S$ is a scaling factor to map $x$ within clipping range $[-\alpha, \alpha]$ into the range of $[-2^{N-1}, 2^{N-1}-1]$, $round(\cdot)$ is the rounding operation and $\hat x$ is the quantized integer value. For most symmetric uniform quantization, $\alpha$ is defined by $\alpha=max(|\bm{x}|)$ to cover all of the values. After quantization, memory footprint and computational cost will be reduced. Then, we can easily obtain the dequantized value $x_d$ as follows:
\begin{equation}
    x_d=\hat x\cdot S \ .
\label{eq:deq}
\end{equation}
Due to the poor representation ability of limited bit width, there exists quantization error between the dequantized value $x_d$ and the original floating-point value $x$, which may involve quantization noise and lead to accuracy loss. 

To recover the quantized model performance, there exist two challenges for DFQ methods: (1) For statistic activation quantization, clipping range of activation values should be determined without access to the training data. (2) To recover the degraded performance, fine-tuning is used to adjust the weights of the quantized models without training data. To solve these challenges, current DFQ methods try to generate synthetic data which are similar to the original training data. 

However, current generative DFQ methods neglected the inter-class separability of semantic features in synthetic data generation. From our perspective, this will be the most critical factor for the performance recovery of quantized model.

\subsection{Empirical Observation}

\begin{figure}[t]
\makeatletter
\makeatother
\centering
\vspace{-5mm}
\subfloat[]{\includegraphics[width=0.48\linewidth]{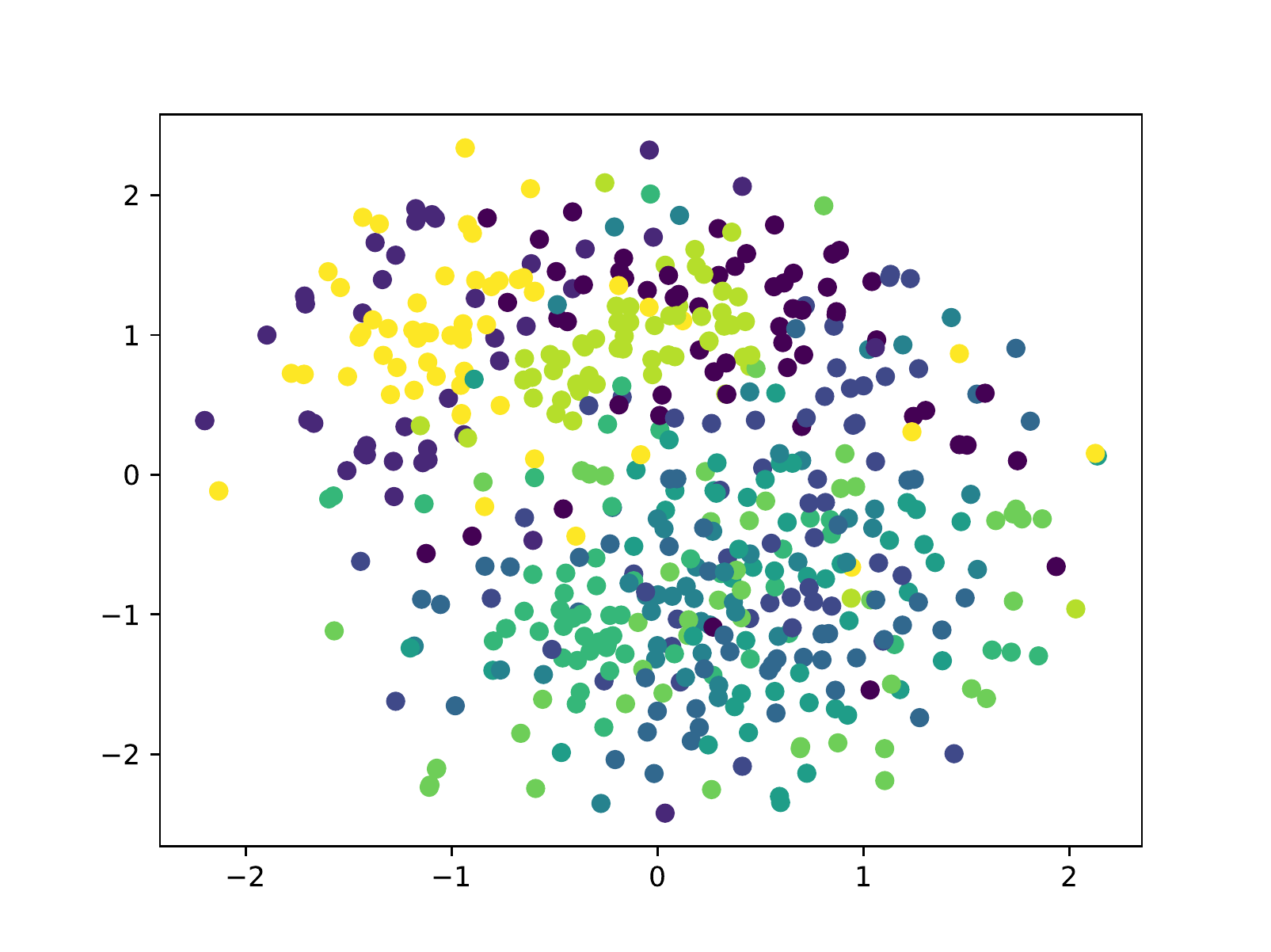}}
\quad
\subfloat[]{\includegraphics[width=0.48\linewidth]{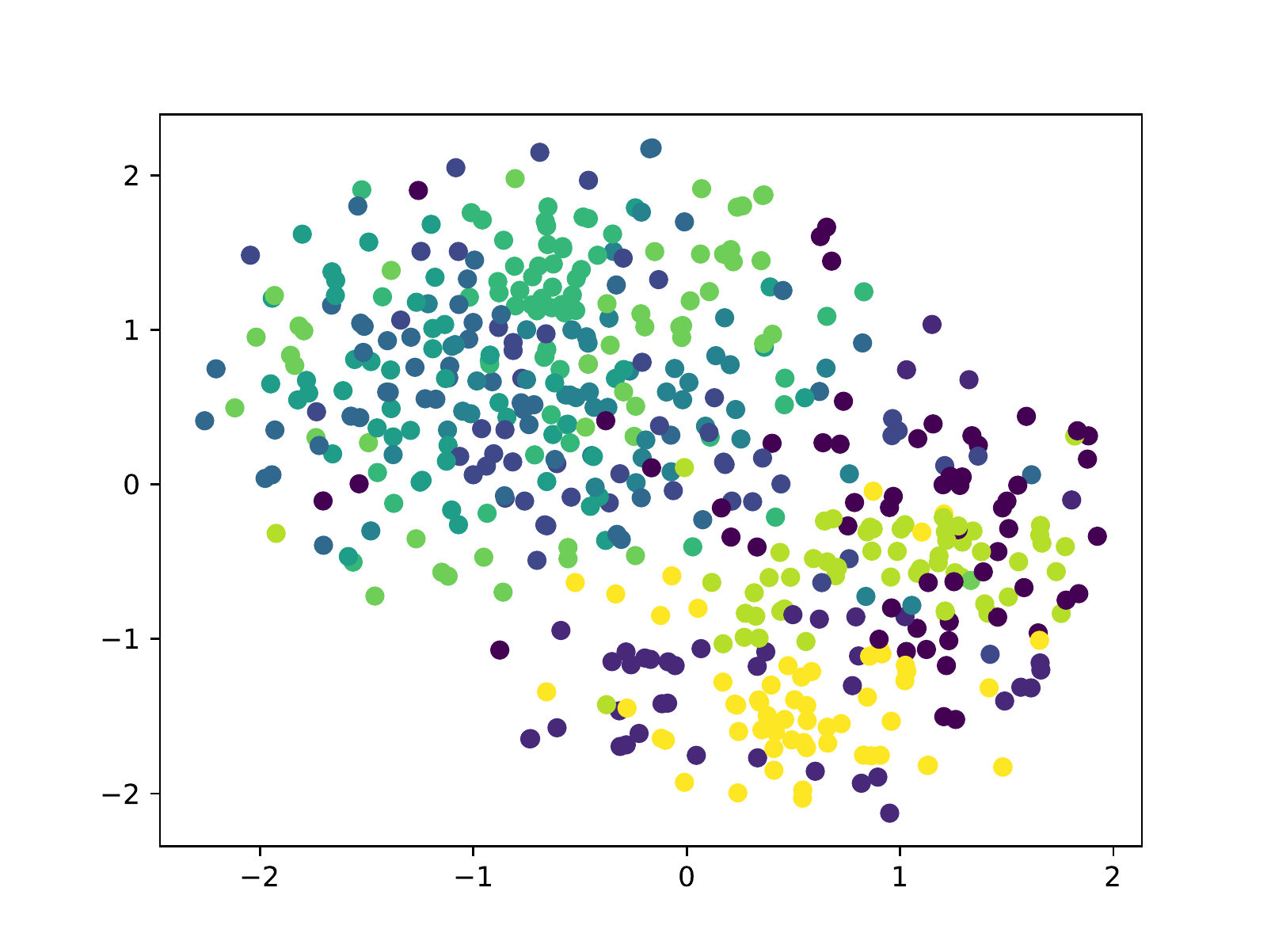}}
\vspace{-2mm}
\subfloat[]{\includegraphics[width=0.48\linewidth]{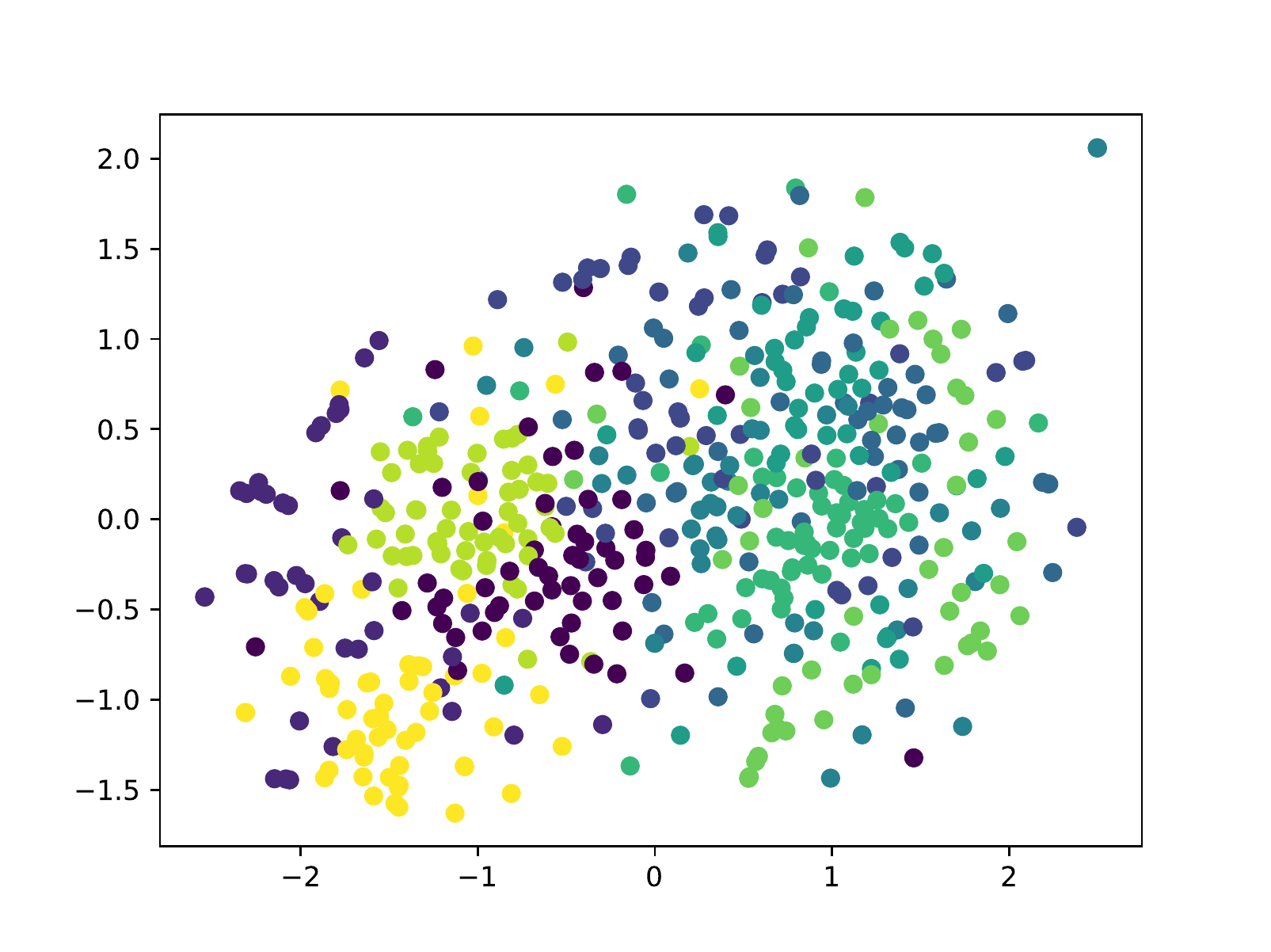}}
\quad
\subfloat[]{\includegraphics[width=0.48\linewidth]{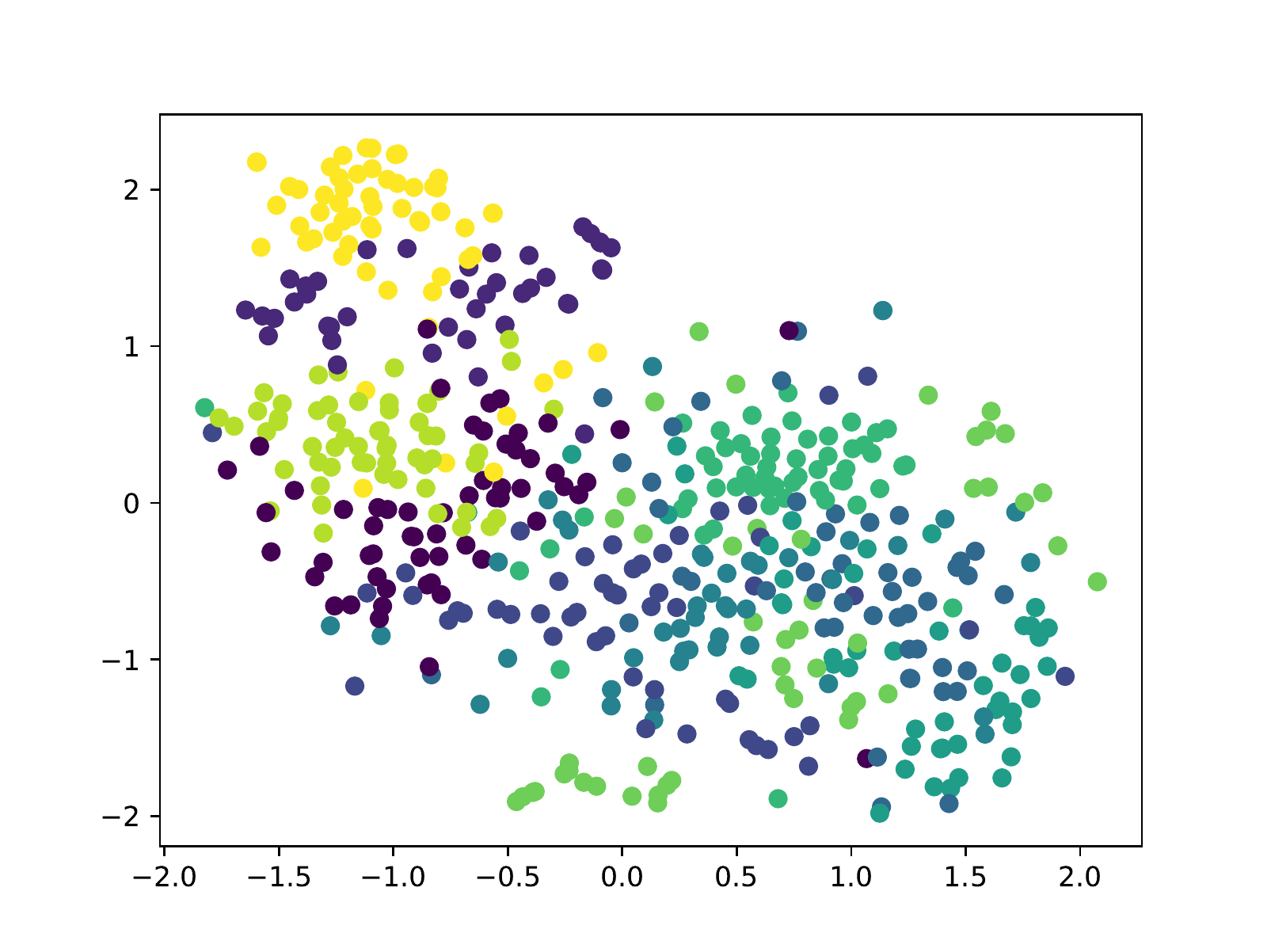}}
\vspace{-2mm}
\subfloat[]{\includegraphics[width=0.48\linewidth]{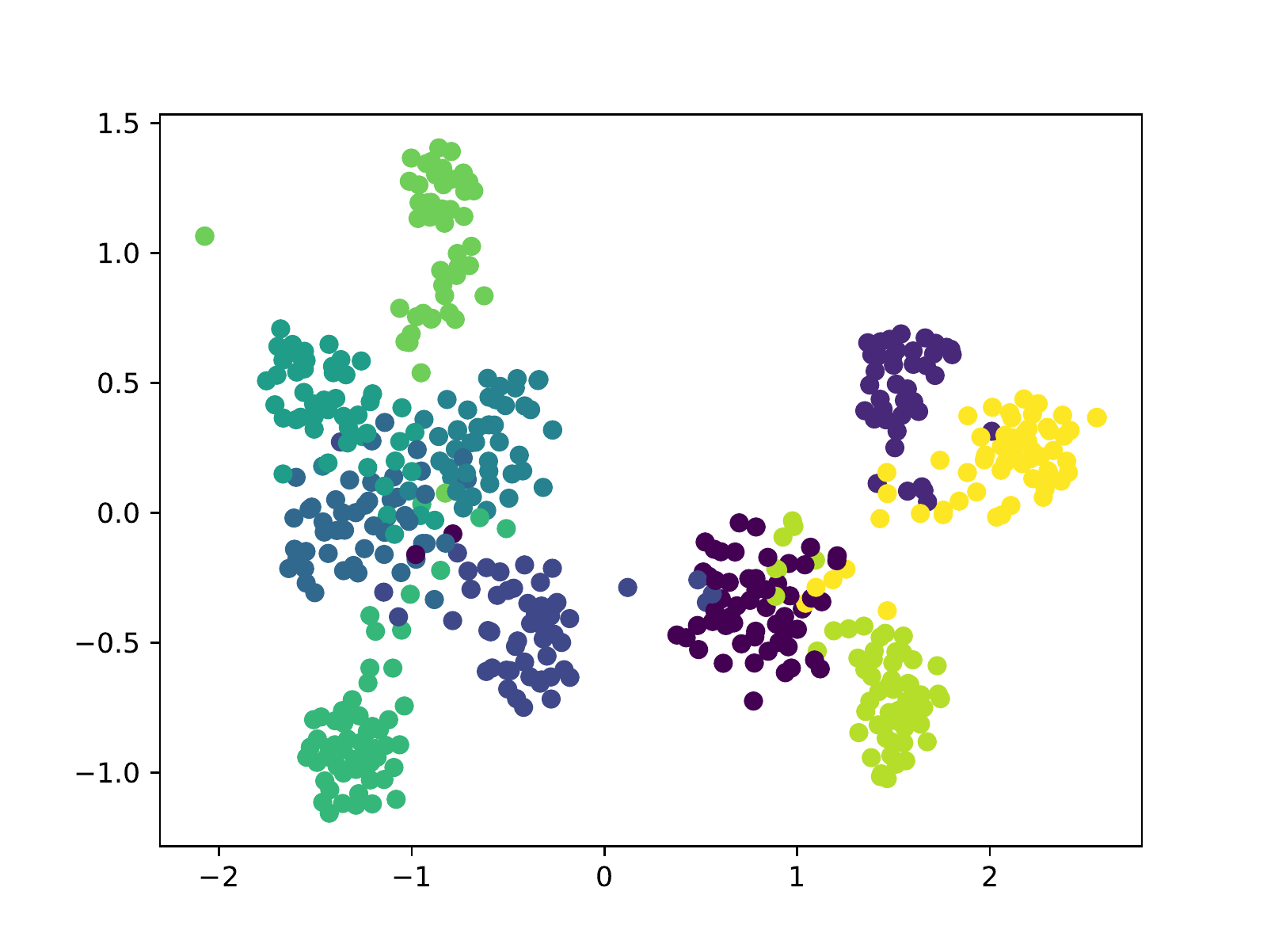}}
\quad
\subfloat[]{\includegraphics[width=0.48\linewidth]{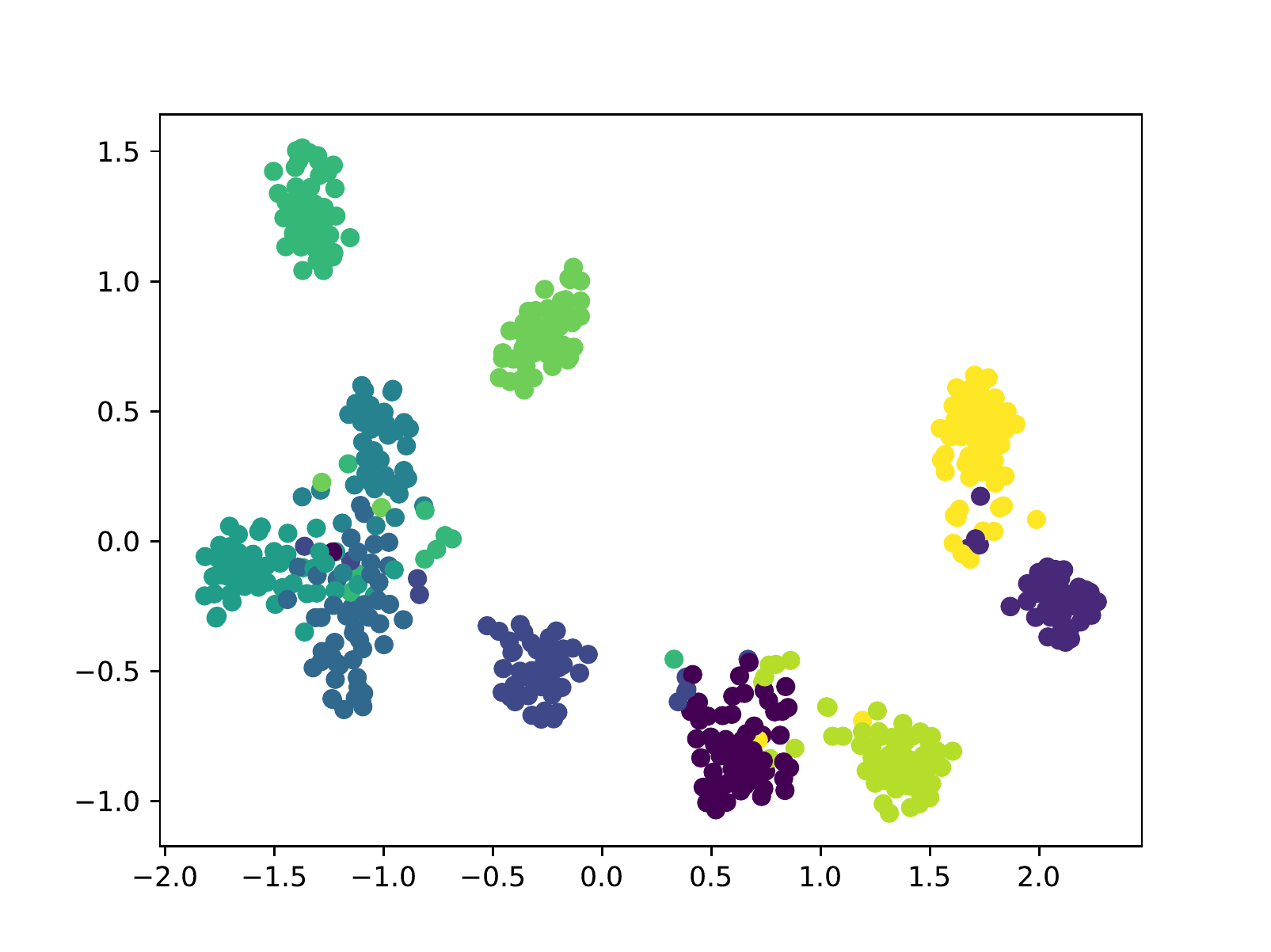}}
\vspace{3mm}
\caption{t-SNE visualization results of the deep layer features calculated by the ResNet-20 model inferring on CIFAR-10 dataset. From (a) to (f) correspond to the features from $14$th layer to $19$th layer. Clearly, the inter-class separability is enhanced when the layer gets deeper.}
\label{fig:tsne}
\vspace{-3mm}
\end{figure}

To highlight our motivation on the inter-class separability of semantic features, we conduct some pilot experiments on the DNN features to observe the dynamic transformation of this separability over different layers. As illustrated in Figure \ref{fig:tsne}, we feed the real images into ResNet-18 and visualize the feature distributions. According to the visual results, we can have the following observations:  
\begin{enumerate}

\item Feature distribution has the property of aggregation over different categories. As shown in Figure \ref{fig:tsne}(e) and \ref{fig:tsne}(f), there exist boundaries over the feature distribution between classes, which will lead to inter-class separability. 

\item As the layer getting deeper, the feature distributions are more separable and can be easily grouped. We can easily distinguish the features of the $18$th and $19$th layers (see Figure \ref{fig:tsne}(e) and \ref{fig:tsne}(f)), while the boundaries of clusters become blurred in the $16$th and $17$th layers (see Figure \ref{fig:tsne}(c) and \ref{fig:tsne}(d)). For more shallow layers (see Figure \ref{fig:tsne}(a) and \ref{fig:tsne}(b)), almost no boundary exists.

\end{enumerate}

\begin{figure*}[t]
\makeatletter
\makeatother
\centering
\includegraphics[width=0.99\linewidth, height=0.325\textheight ]{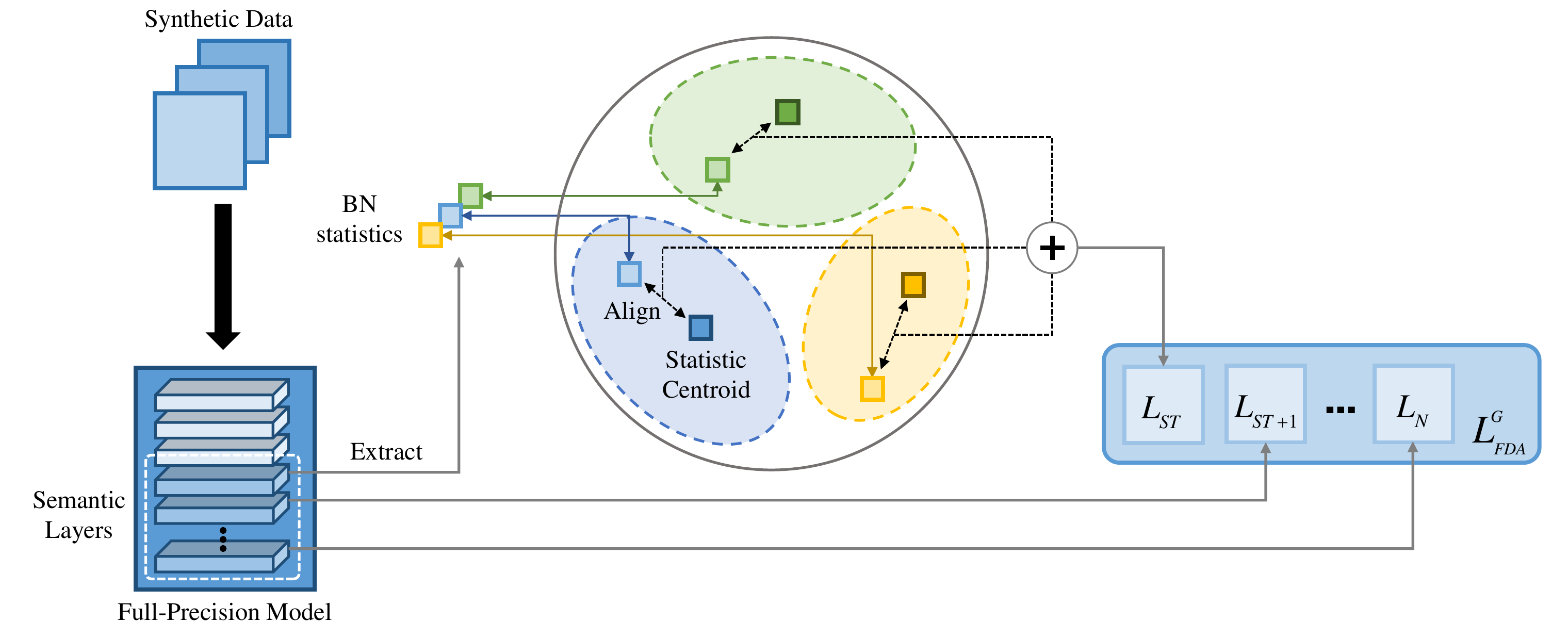}
	\vspace{-2mm}
\caption{The structure of proposed FDA approach. The BN statistics of each class are extracted in deep layers and aligned to the centers corresponding to the classes. The FDA loss is computed by the distance between statistic centers and current statistics. Different colors are used to represent the pseudo labels of centers and statistics. $L_{ST}$ and $L_{N}$ denote the FDA loss of the starting layer and the $N$th layer, respectively.}
\label{fig:sfda}
\vspace{-2mm}
\end{figure*}

\subsection{Proposed Framework}

The observations above indicates that inter-class separability of the semantic features will be one of the hidden properties within the DNN model. As such, we directly utilize the Batch Normalization statistics that save running statistics for feature clustering and alignment.

The structure of the proposed ClusterQ framework is shown in Figure \ref{fig:overview}, which is based on the CGAN mechanism. Specifically, ClusterQ employs the fixed full-precision model $M_{FP}$ as a discriminator. The generator $G$ is trained from scratch to generate synthetic data to fine-tune the quantized model $M_Q$. 

The generator $G$ is trained by loss $L_1(G)$ for classification and global feature distribution retraining. The quantized model $M_Q$ is fine-tuned by loss $L_2(Q)$. More importantly, generator training phase also introduces $L_{DA}^G(G)$ for feature distribution alignment to achieve inter-class separability in semantic layer. To adapt the distribution shift during generator training, we implement the dynamic centroid update by EMA. Moreover, to avoid mode collapse, we introduce the diversity enhancement to eliminate the distribution homogenization. 

\subsection{Feature Distribution Alignment (FDA)}
The structure of FDA is illustrated in Figure \ref{fig:sfda}. In general, we directly utilize the running mean and standard deviation statistics of BN layers, i.e. BN statistics, align to the centroids of different categories. To extract the information of class-wise feature distribution, loss $L^G_{FDA}(G)$ is computed by adding distance between centroids and BN statistics. Note that feature distributions in shallow layers have no aggregation property, we only act alignment on deep layers. In this way, the knowledge of pre-trained model can be extracted to enhance the classification boundaries of quantized model. The FDA process is elaborated as follows: 

\begin{enumerate}
    \item Since we take the pre-trained model as discriminator, the generator $G$ will warm up firstly to produce fake data with high confidence and diversity. 
    
    \item Then, we initialize the centroids for each class in each semantic layer. By feeding Gaussian data with soft labels of each category, the generator will produce a series of synthetic data. During the inference of pre-trained model on these fake data, we can extract the corresponding BN statistics in each semantic layer. 
    
    \item During the generator training phase, we perform feature distribution alignment in each deep layer. Specifically, our FDA loss function $L^G_{FDA}(G)$ adds the Euclidean distance between the running BN statistics and statistic centroid of each class in deep layers as follows: 
    \begin{equation}
        L^G_{FDA}(G)=\sum_{C=0}^{N_C} \sum_{l=l_{st}}^L \big\| \hat{\mu}^C_l - \mu^C_l \big\|^2_2 + \big\| \hat{\sigma}^C_l - \sigma^C_l \big\|^2_2 ,
        \label{eq:SFDA}
    \end{equation}
    where $\hat{\mu}^C_l$ and $\hat{\sigma}^C_l$ denote the running mean and standard deviation for class $C$ at the $l$th layer in the full-precision model, $\mu^C_l$ and $\sigma^C_l$ represent the corresponding mean and standard deviation of the centroids, respectively. $l_{st}$ denotes the starting layer that contains semantic features. And $N_C$ denotes the number of classes. 
\end{enumerate}

In addition, during the generator training, the BN statistics obtained by mis-classified synthetic data will not participate in the computation process of loss $L^G_{FDA}(G)$ . 

Specifically, the FDA process can extract class-wise distribution information and promote the quality of synthetic data. Thus, during the fine-tuning process, the learned classification boundary will be further enhanced. 

\subsection{Centroid Updating}
\label{subsec:cen}

Even though the generator $G$ has warmed up, the initialized statistic centers may not be the optimal solution for FDA process. On one hand, the centers are initialized at the point that alignment has not started. We cannot obtain any distribution information from real data, since it is a unsupervised learning problem in essence. On the other hand, the feature distribution may shift during training, as we cannot constrain the distance between different centers. Initialized statistic centers will limit further alignment to feature distribution. 

For these reasons, we need to update the centroids during generator training to release the negative effects. Thus, we update the centroids by running BN statistics during generator training. 
Specifically, for trading-off between previous and current distribution, we apply exponential moving average (EMA) directly on it to update the centroids as follows: 
\begin{equation}
    \begin{cases} 
    \mu_l^C=(1-\beta_{FD})\mu_l^C + \beta_{FD}\hat{\mu}^C_l\\
    \sigma_l^C=(1-\beta_{FD})\sigma_l^C + \beta_{FD}\hat{\sigma}^C_l
    \end{cases}
    ,
\label{eq:ema}
\end{equation}
where $\hat{\mu}^C_l$ and $\hat{\sigma}^C_l$ denote the running mean and standard deviation corresponding to class $C$, respectively. $\beta_{FD}$ is the decay rate of EMA, which trades off the importance of previous and current BN statistics. Therefore, the distribution centroids can dynamically learn the class-wise feature distribution. We will provide experimental results to demonstrate the performance promotion via centroids updating and analyse the impact of the decay rate value. 

\subsection{Diversity Enhancement}

\begin{figure*}[t]
\makeatletter
\makeatother
\centering
\includegraphics[width=0.94\linewidth]{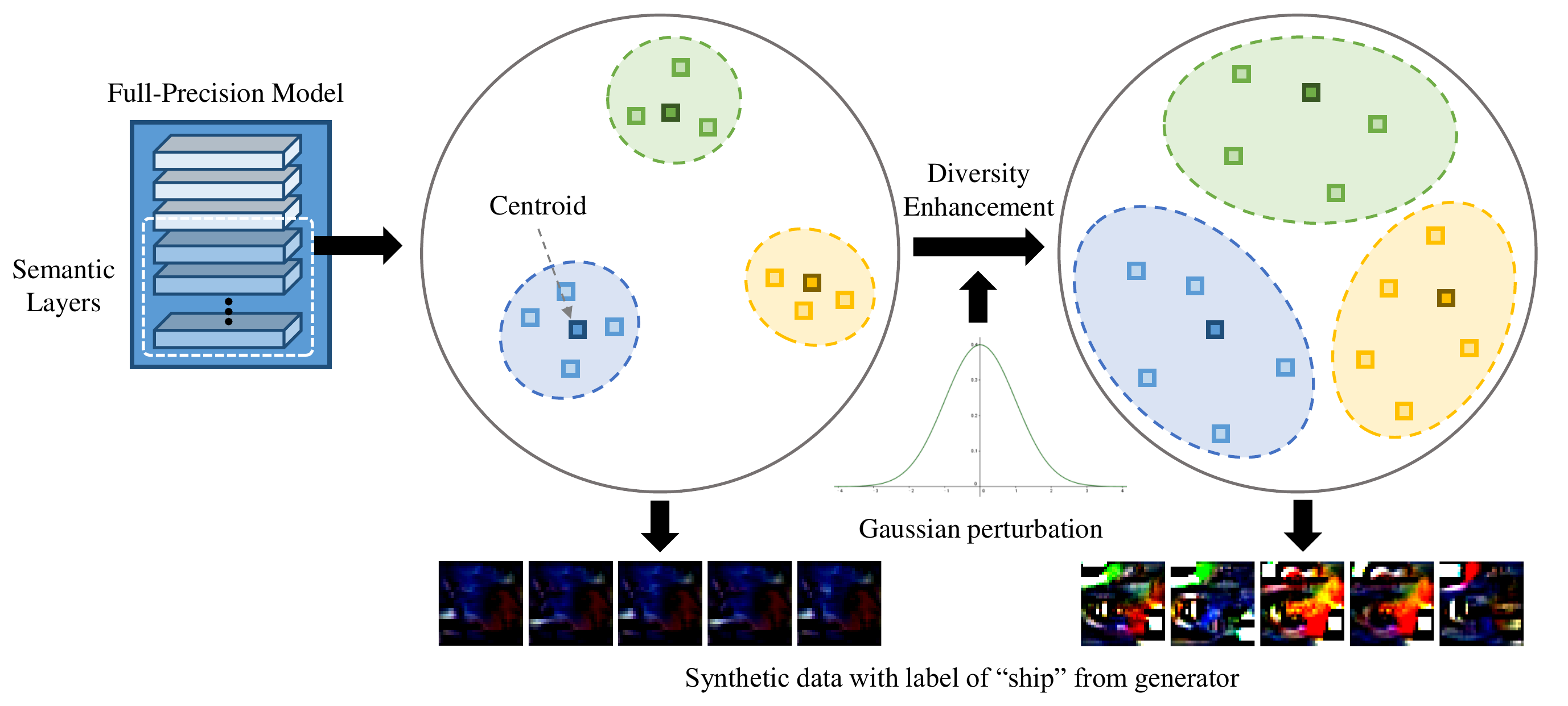}
	\vspace{-2mm}
\caption{Illustration of diversity enhancement process. Based on involving the uniform perturbation into generator training, the feature distribution is allowed to shift around the centers. As a result, class-wise mode collapse is released in data generation. }
\label{fig:icv}
	\vspace{-2mm}
\end{figure*}

Although our proposed FDA method can obtain high inter-class separability of semantic features, the distribution alignment may also cause vulnerability of class-wise mode collapse which will also degrade the generalization performance of quantized model. That is, the distribution of real data cannot be covered by the synthetic data. For example, given Gaussian input, some generators produce data in fixed mode. 

To expand mode coverage, we employ diversity enhancement approach to shift the BN statistic distribution around the centroids. Specifically, we introduce the perturbation of Gaussian distribution to release the feature distribution homogenization of each class, which is caused by feature distribution over-fitting \cite{zhang2021diversifying}. To solve this issue, we define the diversity enhancement loss $L_{DE}$ as follows: 
\begin{equation}
\begin{aligned}
    L_{DE}(G)=\sum_{C=0}^{N_C}\sum_{l=l_{st}}^L &\big\| \hat{\mu}^C_l - \mathcal{N}(\mu^C_l,\lambda_{\mu}) \big\|^2_2 \\+ &\big\| \hat{\sigma}^C_l-\mathcal{N}(\sigma_l^C,\lambda_\sigma)\big\|^2_2\ ,
\end{aligned}
\label{eq:LICV}
\end{equation}
where $\mathcal{N}(\cdot,\cdot)$ denotes Gaussian noise, $\lambda_{\mu}$ and $\lambda_{\sigma}$ denote the distortion levels of diversity enhancement. In this way, we can allow the running mean $\hat{\mu}^C_l$ and standard deviation $\hat{\sigma}^C_l$ for each class $C$ to shift within a dynamic range around the centroids $\mu^C_l$ and $\sigma_l^C$ respectively. As shown in Figure \ref{fig:icv}, semantic feature distribution space cannot be covered without diversity enhancement.In contrast, homogenization can be released with the introduction of perturbation to FDA process. Experiments have verified the effect of diversity enhancement loss $L_{DE}$ to mitigate the mode collapse in synthetic data generation. 

\subsection{Training Process}

In this subsection, we summarize the whole training process for comprehensive understanding to our ClusterQ. Based on the quantized model $M_Q$ processed by Eq.(\ref{eq:quant}) and the full-precision model $M_{FP}$ as discriminator, our scheme trains the generator $G$ to produce synthetic data and fine-tunes the parameters of the quantized model $M_Q$ alternately. Note that our implementation is based on the framework of GDFQ \cite{xu2020generative}. 

At the beginning of the generator $G$ training, i.e., warm-up process, we fix the parameters of quantized model $M_Q$ to avoid being updated, because the generated synthetic data lack of diversity and textures. And note that full-precision model $M_{FP}$ are fixed during the whole process. For global BN statistic matching and classification, the loss function $L_1(G)$ is denoted as follows: 
\begin{equation}
    L_1(G)=L^G_{CE}(G)+\alpha_1 L^G_{BNS}(G) ,
\label{eq:L1}
\end{equation}
where $\alpha_1$ is a trade-off parameter. The term $L^G_{CE}(G)$ utilizes the cross-entropy loss function $CE(\cdot,\cdot)$. The term $L^G_{BNS}(G)$ denotes the loss to match global BN statistics in each layer. 

After finishing the warm-up process, we utilize the synthetic data to fine-tune the quantized model, and initialize the BN statistic centroids. Then, the FDA loss $L^G_{FDA}(G)$ and the diversity enhancement loss $L^{G}_{DE}(G)$ will be involved into the generator training, which is formulated as
\begin{equation}
\begin{aligned}
    L^{'}_1(G)=L_1(G)+\alpha_2 L^G_{FDA}(G)+\alpha_3 L^{G}_{DE}(G),
\label{eq:L2}
\end{aligned}
\end{equation}
where $\alpha_2$ and $\alpha_3$ are the trade-off parameters. After that, the centroids will be updated with current BN statistics by EMA.

During the quantized model $M_Q$ fine-tuning phase, we use the following loss function $L_2(M_Q)$:
\begin{equation}
    L_2(M_Q)=L^Q_{CE}(M_Q)+\gamma L^Q_{KD}(M_Q),
\label{eq:L3}
\end{equation}
where $\gamma$ is a trade-off parameter. With the synthetic data and corresponding pseudo label $y$, the parameters of the quantized model are updated by the cross-entropy loss term $L^Q_{CE}(M_Q)$. And the knowledge distillation loss function $L_{KD}^Q(M_Q)$ via Kullback-Leibler divergence loss $KLD(\cdot,\cdot)$ is employed to compare the similarity of output distribution between quantized model $M_Q$ and full-precision model $M_{FP}$.

\section{Experiments}
\label{sec:exp}

\subsection{Experimental Setting}

We compare each method on several popular datasets, including CIFAR10, CIFAR100 \cite{krizhevsky2009learning} and ImageNet (ILSVRC12) \cite{deng2009imagenet}. With $60$ thousand images of pixels $32\times32$, CIFAR10 and CIFAR100 datasets contain $10$ categories for classification. ImageNet has $1000$ categories for classification with $1.2$ million training images and $150$ thousand images for validation. 

For experiments, we perform quantization on ResNet-18 \cite{he2016deep}, MobileNet-V2 \cite{sandler2018mobilenetv2} on ImageNet, and also ResNet-20 on CIFAR10 and CIFAR100. All experiments are conducted on an NVIDIA RTX 2080Ti GPU with PyTorch \cite{paszke2017automatic}. Note that all of the pre-trained model implementations and weights are provided by Pytorchcv\footnote{Computer vision models on PyTorch:
\url{https://pypi.org/project/pytorchcv/}}. Our codes are available at \href{https://github.com/DiamondSheep/ClusterQ}{https://github.com/DiamondSheep/ClusterQ}.

For implementation, we follow some hyperparameter settings of GDFQ \cite{xu2020generative}. We set $400$ epochs for the generator training, $200$ epochs for quantized model fine-tuning and $50$ epochs for generator warm-up. And for trade-off parameters, we set $0.1$ for $\alpha_1$, $0.9$ for $\alpha_2$, $0.6$ for $\alpha_3$ and $1.0$ for $\gamma$. For EMA, we set the decay rate $\beta_{FD}$ to $0.2$. In $L_{DE}$, the uniform distribution parameters $\lambda_p$ are set to $0.3$ and $0.15$ for mean and standard deviation, respectively. 

\subsection{Comparison Results}

\begin{table}[t]
	\footnotesize
	\centering
	\caption{Comparison results on ImageNet dataset.} 
		\vspace{-1mm}
	\setlength{\tabcolsep}{1.6mm}{
		\begin{tabular}{cccc}
			\hline
			DNN Model  & Precision & Quantization Method & Top1 Accuracy \\
			\hline
			\multirow{15}*{ResNet-18}   & W32A32 & Baseline & 71.470\% \\
			\cline{2-4}
			& \multirow{7}*{W4A4} & ZeroQ  & 20.770\% \\
			&                     & GDFQ   & 60.704\% \\
			&                     & DSG    & 34.530\% \\
			&                     & Qimera & 63.840\% \\
			&                     & AutoReCon & 61.600\% \\
			&                     & DDAQ   & 58.440\% \\
			&                     & \textbf{Ours} & \textbf{64.390\%} \\
			\cline{2-4}
			& \multirow{4}*{W4A8} & ZeroQ$^\dagger$ & 51.176\% \\
			&                     & GDFQ$^\dagger$ & 64.810\% \\
			&                     & Qimera$^\dagger$ & 65.784\% \\
			&                     & \textbf{Ours} & \textbf{67.826\%} \\
			\cline{2-4}
			& \multirow{3}*{W8A8} & GDFQ$^\dagger$ & 70.788\% \\
			&                     & Qimera$^\dagger$ & 70.664\% \\
			&                     & \textbf{Ours} & \textbf{70.838\%} \\
			\hline
			\multirow{14}*{MobileNet-V2} & W32A32 & Baseline & 73.084\% \\
			\cline{2-4}
			& \multirow{6}*{W4A4} & ZeroQ  & 10.990\% \\
			&                     & GDFQ   & 59.404\% \\
			&                     & Qimera & 61.620\% \\
			&                     & AutoReCon & 60.020\% \\
			&                     & DDAQ & 52.990\% \\
			&                     & \textbf{Ours} & \textbf{63.328\%} \\
			\cline{2-4}
			& \multirow{4}*{W4A8} & ZeroQ$^\dagger$ & 13.955\%  \\
			&                     & GDFQ$^\dagger$ & 64.402\%   \\
			&                     & Qimera$^\dagger$ & 66.486\%  \\
			&                     & \textbf{Ours} & \textbf{68.200\%} \\
			\cline{2-4}
			& \multirow{3}*{W8A8} & GDFQ$^\dagger$ & 72.814\% \\
			&                     & Qimera$^\dagger$ & 72.772\%  \\
			&                     & \textbf{Ours} & \textbf{72.820\%} \\
			\hline
	\end{tabular}}
	\label{tab:imagenet}
\end{table}

%
\begin{table}[t]
	\footnotesize
	\centering
	\caption{Comparison results on CIFAR100 dataset.}
		\vspace{-1mm}
	\setlength{\tabcolsep}{1.5mm}{
		\begin{tabular}{cccc}
			\hline
			DNN Model  & Precision & Quantization Method & Top1 Accuracy \\
			\hline
			\multirow{13}*{ResNet-20} & W32A32                & Baseline & 70.33\%      \\
			\cline{2-4}
			& \multirow{4}*{W4A4}   & ZeroQ  & 45.20\%        \\
			&                       & GDFQ   & 63.91\%        \\
			&                       & Qimera & 65.10\%        \\
			&                       & \textbf{Ours} & \textbf{67.09\%} \\
			\cline{2-4}
			& \multirow{4}*{W4A8}   & ZeroQ$^\dagger$ & 58.606\% \\
			&                       & GDFQ$^\dagger$ & 67.33\%         \\
			&                       & Qimera$^\dagger$ & 68.89\%         \\
			&                       & \textbf{Ours} & \textbf{69.68\%} \\
			\cline{2-4}
			& \multirow{4}*{W8A8}   & ZeroQ$^\dagger$ & 70.128\% \\
			&                       & GDFQ$^\dagger$ & 70.39\%           \\
			&                       & Qimera$^\dagger$ & 70.40\%          \\
			&                       & \textbf{Ours} & \textbf{70.43\%} \\
			\hline
	\end{tabular}}
	\label{tab:cifar100}
	\vspace{-3mm}
\end{table}

%
\begin{table}[t]
	\footnotesize
	\centering
	\vspace{-3mm}
	\caption{Comparison results on CIFAR10 dataset.}
	\setlength{\tabcolsep}{1.5mm}{
		\begin{tabular}{cccc}
			\hline
			DNN Model  & Precision & Quantization Method & Top1 Accuracy \\
			\hline
			\multirow{13}*{ResNet-20} & W32A32                & Baseline & 93.89\%      \\
			\cline{2-4}
			& \multirow{4}*{W4A4}   & ZeroQ  & 73.53\%        \\
			&                       & GDFQ   & 86.23\%        \\
			&                       & Qimera & 91.23\%        \\
			&                       & \textbf{Ours} & \textbf{92.06\%} \\
			\cline{2-4}
			& \multirow{4}*{W4A8}   & ZeroQ$^\dagger$ & 90.845\% \\
			&                       & GDFQ$^\dagger$ & 93.74\%         \\
			&                       & Qimera$^\dagger$ & 93.63\%         \\
			&                       & \textbf{Ours} & \textbf{93.84\%} \\
			\cline{2-4}
			& \multirow{4}*{W8A8}   & ZeroQ$^\dagger$ & 93.94\% \\
			&                       & GDFQ$^\dagger$ & 93.98\%           \\
			&                       & Qimera$^\dagger$ & 93.93\%          \\
			&                       & \textbf{Ours}$^\dagger$ & \textbf{94.07\%} \\
			\hline
	\end{tabular}}
	\label{tab:cifar10}
\end{table}

To demonstrate the performance of our ClusterQ, we compare it with several closely-related methods, i.e., ZeroQ \cite{cai2020zeroq}, GDFQ \cite{xu2020generative}, Qimera \cite{choi2021qimera}, DSG \cite{zhang2021diversifying}, DDAQ \cite{li2022dual} and AutoReCon \cite{zhu2021autorecon}. The comparison results based on ImageNet, CIFAR100 and CIFAR10 are described in Tables \ref{tab:imagenet}, \ref{tab:cifar100} and \ref{tab:cifar10}, respectively. Note that W$n$A$m$ stands for the quantization bit-width with $n$-bit weight and $m$-bit activation. The baseline with W32A32 denotes the full-precision model accuracy. The character $^\dagger$ means that the result is obtained by ourselves. By considering the practical applications, we also conduct quantization experiments with different precision settings. Moreover, we choose the bit number with power of two in all experiments for facilitating the deployment. 

\subsubsection{Results on ImageNet} 
As can be seen in Table \ref{tab:imagenet}, with the same precision setting based on the ResNet-18 and MobileNet-V2, our method performs better than its competitors. Specifically, our method performs beyond the most closely-related GDFQ method a lot, especially for the case of lower precision. By comparing with the current state-of-the-art method Qimera, our method still outperforms it $1.708\%$ for MobileNet-V2 that is, in fact, more difficult to be compressed due to smaller weights. One can also note that, with the reduction of precision bits, the presentation ability of the quantized value becomes limited and leads to more performance degradation. In this case, our ClusterQ retains the performance of quantized model better than other compared competitors. 

\subsubsection{Results on CIFAR10 and CIFAR100}
From the results in Tables \ref{tab:cifar100} and \ref{tab:cifar10} based on ResNet-20, similar conclusions can be obtained. That is, our method surpasses the current state-of-the-art methods in terms of accuracy loss in this investigated case. In other words, the generalization performance of our method on different models and datasets can be verified. 

\subsection{Visual Analysis}

\begin{figure}[t]
    \makeatletter
    \makeatother
    \centering
    \includegraphics[width=0.92\linewidth]{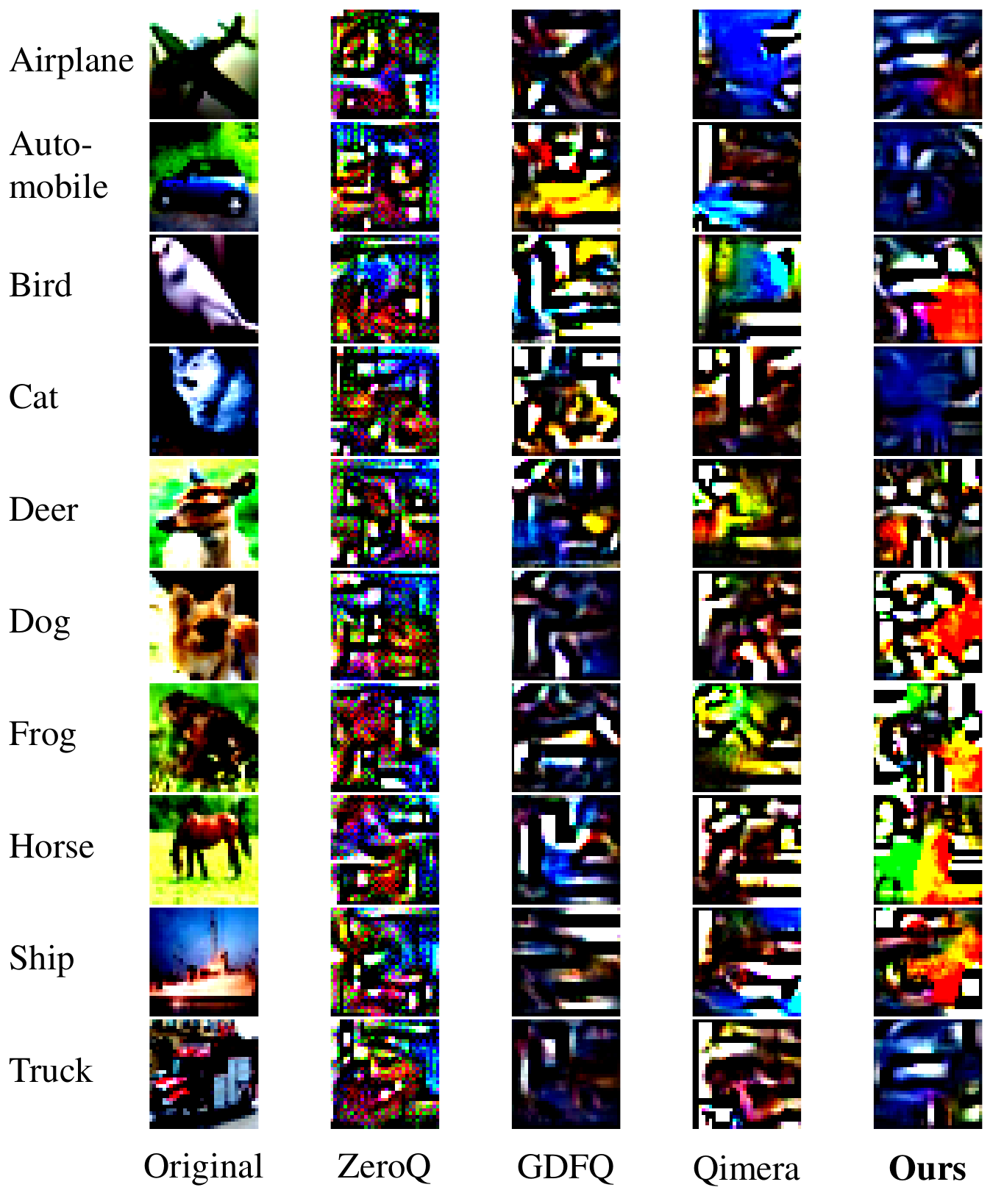}
    	\vspace{-3mm}
    \caption{Synthetic data generated by generative DFQ methods with the pre-trained ResNet-20 model on CIFAR10 dataset. Each row corresponds to the category, except for ZeroQ, since it generates data without soft labels.}
    \label{fig:visual_1}
        	\vspace{-3mm}
\end{figure}

\begin{figure}[t]
    \makeatletter
    \makeatother
    \centering
    \includegraphics[width=0.84\linewidth]{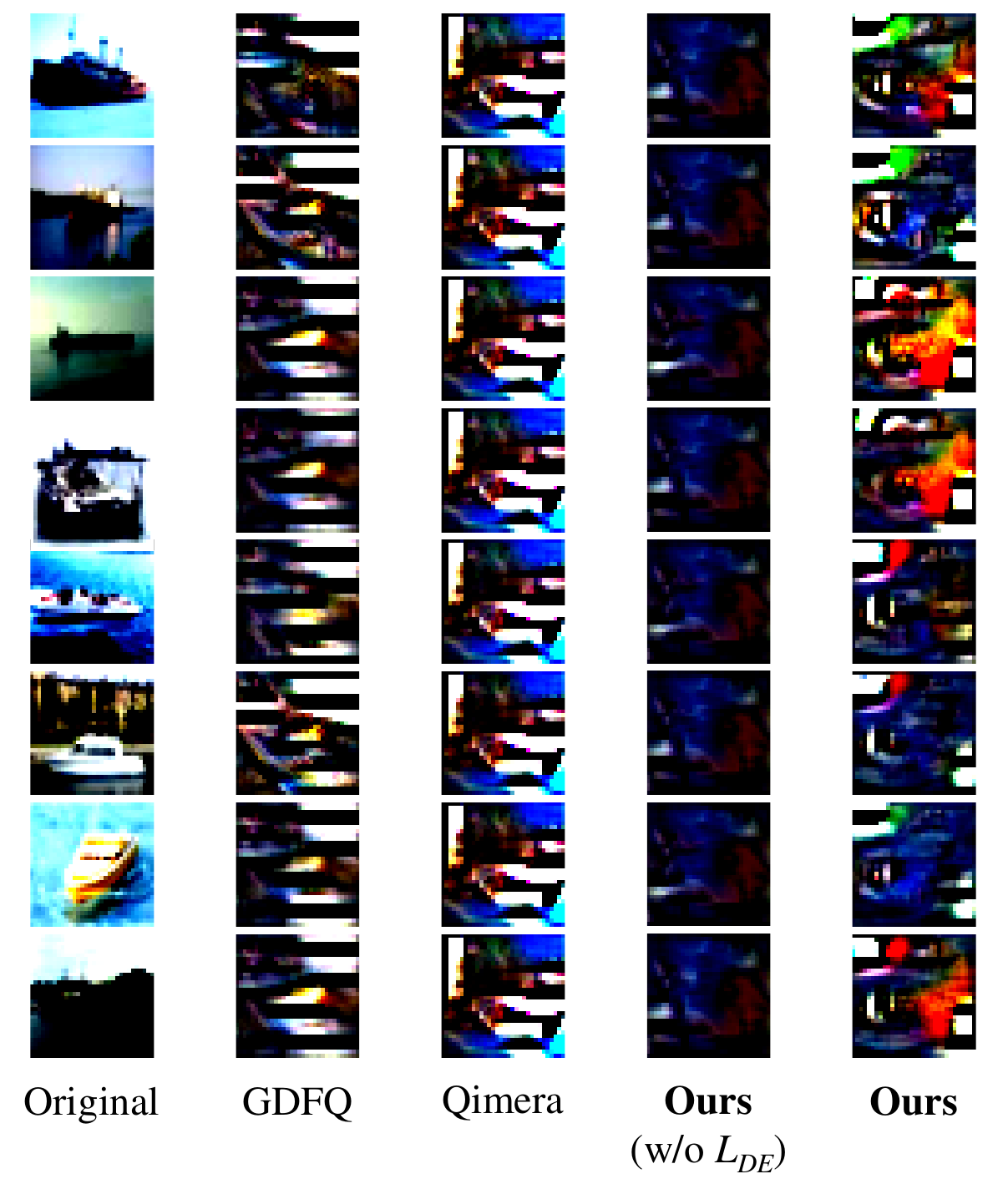}

    \vspace{-4mm}
    \caption{Visualization comparision of the synthetic data produced by DFQ methods with the pre-trained ResNet-20 model on CIFAR10, where "ship" is chosen as an example. ``w/o $L_{DE}$'' denotes the results without $L_{DE}$. }
    \label{fig:visual_2}
\end{figure}

In addition to the above numerical results, we also would like to perform the visual analysis on the generated synthetic data, which will directly impact the performance recovery of each quantized model. In Figure \ref{fig:visual_1}, we visualize the synthetic data with labels generated by existing methods (i.e., ZeroQ, GDFQ and Qimera) based on the ResNet-20 over CIFAR10. We select the synthetic data with label "ship" as an example and show the results in Figure \ref{fig:visual_2}.

\begin{table}[t]
\footnotesize
\centering
\vspace{-2.5mm}
\caption{Ablation studies on ResNet-18 over ImageNet.}
\setlength{\tabcolsep}{5mm}{
\begin{tabular}{cccc}
\hline
Model  & $L_{DE}$ & EMA & Top1 \\
\hline
\multirow{4}*{ResNet-18} & $\surd$ & $\surd$ & 64.390\%  \\
                          & $\surd$ & - & 63.646\%  \\
                          & - & $\surd$ & 63.590\%  \\
                          & - & - & 63.068\%  \\
\hline
\end{tabular}
\label{tab:ablation}}
\vspace{-2.5mm}
\end{table}

\begin{figure}[t]
	\makeatletter
	\makeatother
	\vspace{2.5mm}
	\centering
	\includegraphics[width=0.99\linewidth]{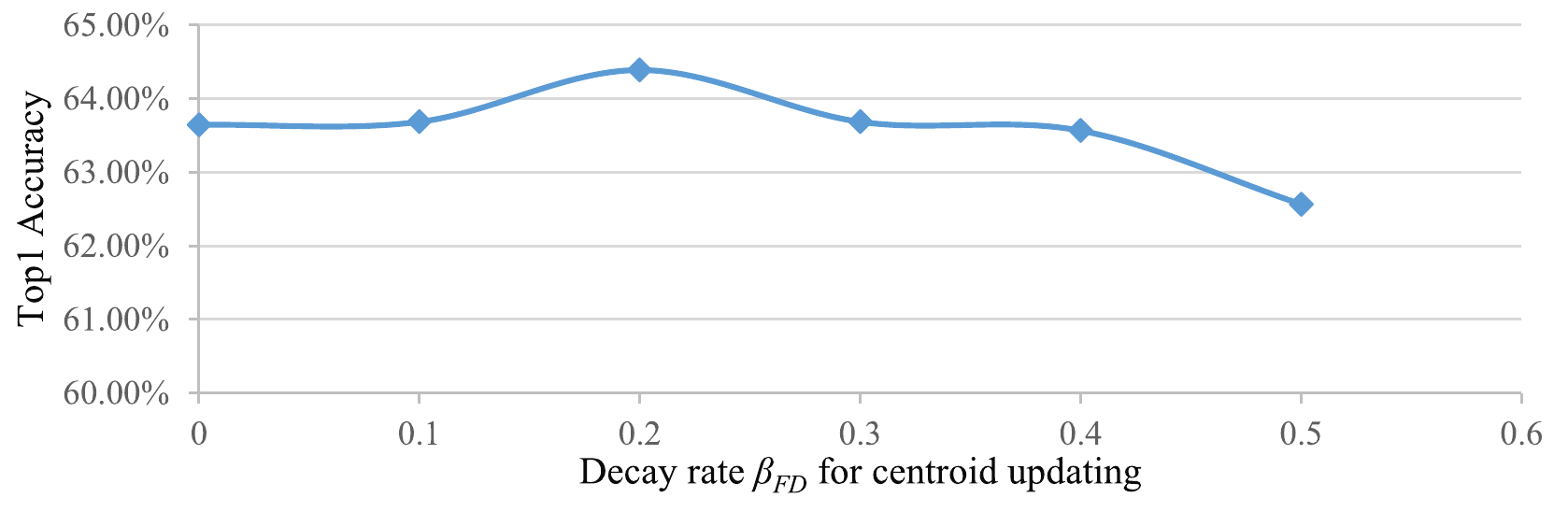}
	
	\vspace{-2mm}
	\caption{Sensitivity analysis of the decay rate of EMA for centroid updating. We conduct the experiments by quantizing ResNet-18 on ImageNet dataset. The quantized model performs the best at the point of $\beta_{FD}=0.2$.}
	\label{fig:decay}
\end{figure}

\begin{figure}[t]
	\makeatletter
	\makeatother
	\centering
	\includegraphics[width=0.99\linewidth]{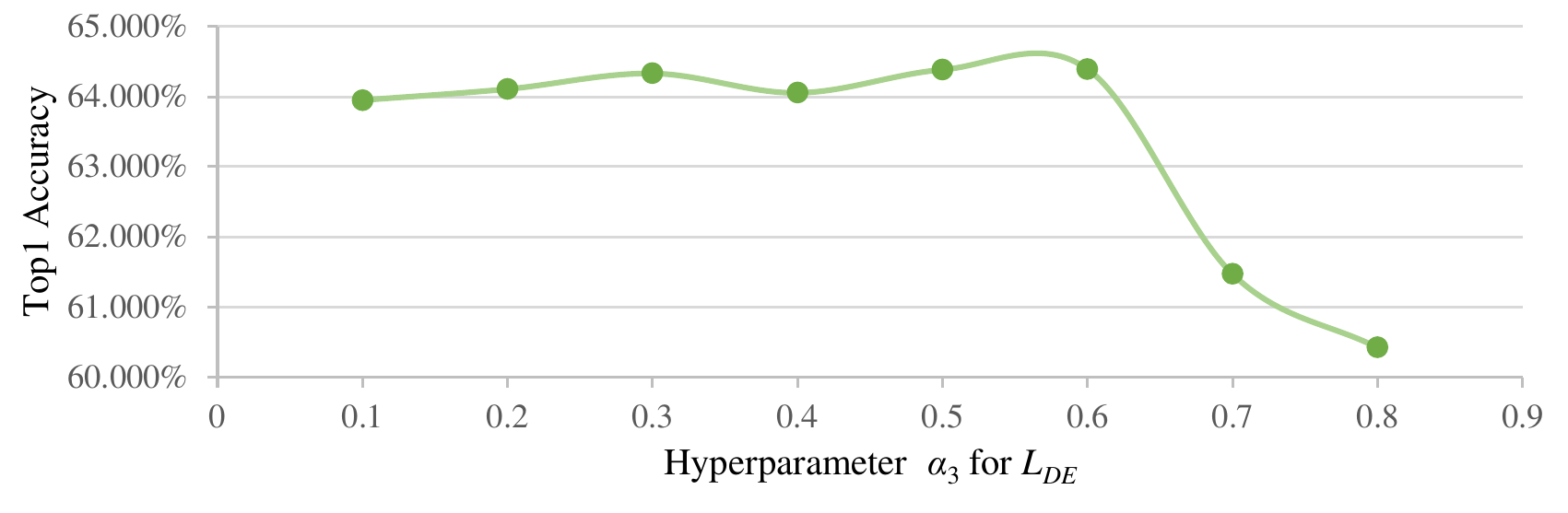}
	
	\vspace{-2mm}
	\caption{Sensitivity analysis of $\alpha_3$ for diversity enhancement. We conduct the experiments by quantizing ResNet-18 on ImageNet dataset. As $\alpha_3$ goes up to $0.6$, the performance of quantized model will increase. However, the performance of the quantized model falls down while $\alpha_3$ goes above $0.6$.}
	\label{fig:sensicv}
	\vspace{-2.5mm}
\end{figure}

As shown in Figure \ref{fig:visual_1}, without learning classification boundaries, the data generated by ZeroQ have less class-wise discrepancy. For GDFQ, the generated data can be distinguished into different classes, but containing less detailed textures. Based on FDA, our ClusterQ can produce the synthetic data with more useful information. With abundant color and texture, the data generated by Qimera are similar to that of ours. However, as shown in Figure.\ref{fig:visual_2}, the little variance of the images within each class indicates that they encounter class-wise mode collapse. In contrast, by simultaneously considering the diversity enhancement, the generated synthetic data of the same class by ClusterQ can maintain variety on color, texture and structure. To show the effect of diversity enhancement, we also visualize the synthetic data produced by ClusterQ without $L_{DE}$ in Figure.\ref{fig:visual_2}, which lead to class-wise mode collapse. 

\subsection{Ablation Studies}

We first evaluate the effectiveness of each component in our ClusterQ, i.e., diversity enhancement and EMA. We conduct experiments to quantize the ResNet-18 into W4A4 on ImageNet dataset, and describe the results in Table \ref{tab:ablation}. We see that without the diversity enhancement or EMA, the performance improvement of quantized model is limited. That is, both diversity enhancement or EMA are important for our method. 

Then, we also analyze the sensitivity of our method to the decay rate $\beta_{FD}$ in Figure \ref{fig:decay}. According to \ref{subsec:cen}, we set the decay rate $\beta_{FD}$ to control the centroid updating and trade. It is clear that the quantized model achieves the best result, when $\beta_{FD}$ equals to $0.2$. The performance is reduced when the decay rate is lower than $0.2$, since in such cases the centroids cannot adapt to the distribution changing. Moreover, if $\beta_{FD}$ is increased beyond $0.2$, the centroids will fluctuate and lead to performance degradation.

In addition, we conduct experiments with different settings of $\alpha_3$ to explore the effect of it. As shown in Figure \ref{fig:sensicv}, when $\alpha_3$ goes up to $0.6$, the performance of quantized model will increase. It demonstrates that diversity enhancement can improve the quality of synthetic data and lead to performance promotion. However, the performance of quantized model falls down when $\alpha_3$ goes above $0.6$, due to the excess distortion which disturbs the classification boundary.

\section{Conclusion}
\label{sec:conc}

We have explored the issue of alleviating the performance degradation when quantizing a model, by enhancing the inter-class separability of semantic features. Technically, a new and effective data-free quantization method referred to as ClusterQ is proposed. The setting of ClusterQ presents a new semantic feature distribution alignment for synthetic data generation, which can obtain high class-wise separability and enhance the diversity of the generated synthetic data. For further improvement, we also incorporate the ideas of diversity enhancement and exponential moving average. Extensive experiments on different deep models over several datasets demonstrate that our method achieves better performance among current data-free quantization methods. In future work, we will focus on exploring how to extend our ClusterQ to other vision tasks. 

\section{Acknowledgments}

This work is partially supported by the National Natural Science Foundation of China (62072151, 61932009, 61822701, 62036010, 72004174), and the Anhui Provincial Natural Science Fund for Distinguished Young Scholars (2008085J30). Zhao Zhang is the corresponding author of this paper, and Jicong Fan is the co-corresponding author. 

\bibliographystyle{IEEEtran.bst}
\balance
\bibliography{reference.bib}

\end{document}